%% file: typeinst.tex
\documentclass[runningheads,a4paper]{llncs}

\usepackage{amssymb}
\usepackage{graphicx}

\usepackage{url}
\urldef{\mailsa}\path|{alfred.hofmann, ursula.barth, ingrid.haas, frank.holzwarth,|
\urldef{\mailsb}\path|anna.kramer, leonie.kunz, christine.reiss, nicole.sator,|
\urldef{\mailsc}\path|erika.siebert-cole, peter.strasser, lncs}@springer.com|

\usepackage{algorithm}% http://ctan.org/pkg/algorithm
\usepackage{algpseudocode}% http://ctan.org/pkg/algorithmicx

\newcommand{\argmax}{\mathop{\rm arg~max}\limits}

\begin{document}

\mainmatter  % start of an individual contribution

\title{Adapting Improved Upper Confidence Bounds for Monte-Carlo Tree Search}

%\titlerunning{Lecture Notes in Computer Science: Authors' Instructions}

\author{Yun-Ching Liu \and Yoshimasa Tsuruoka}

\institute{Department of Electrical Engineering and Information Systems\\ University of Tokyo, Tokyo, Japan\\}


%\toctitle{Lecture Notes in Computer Science}
%\tocauthor{Authors' Instructions}
\maketitle

\begin{abstract}
The UCT algorithm, which combines the UCB algorithm and Monte-Carlo Tree Search (MCTS), is currently the most widely used variant of MCTS. Recently, a number of investigations into applying other bandit algorithms to MCTS have produced interesting results. In this research, we will investigate the possibility of combining the improved UCB algorithm, proposed by Auer et al. \cite{iucb}, with MCTS. However, various characteristics and properties of the improved UCB algorithm may not be ideal for a direct application to MCTS. Therefore, some modifications were made to the improved UCB algorithm, making it more suitable for the task of game tree search. The Mi-UCT algorithm is the application of the modified UCB algorithm applied to trees. The performance of Mi-UCT is demonstrated on the games of $9\times 9$ Go and $9\times 9$ NoGo, and has shown to outperform the plain UCT algorithm when only a small number of playouts are given, and rougly on the same level when more playouts are available. 
\end{abstract}

\section{Introduction}

The development of Monte-Carlo Tree Search (MCTS) has made significant impact on various fields of computer game play, especially the field of computer Go \cite{survey}. The UCT algorithm \cite{uct} is an MCTS algorithm that combines the UCB algorithm \cite{ucb} and MCTS, by treating each node as a single instance of the multi-armed bandit problem. The UCT algorithm is one of the most prominent variants of the Monte-Carlo Tree Search \cite{survey}. 

Recently, various investigations have been carried out on exploring the possibility of applying other bandit algorithms to MCTS. The application of simple regret minizing bandit algorithms has shown the potential to overcome some weaknesses of the UCT algorithm \cite{simple}. The sequential halving on trees (SHOT) \cite{shot} applies the sequential halving algorithm \cite{sh} to MCTS. The SHOT algorithm has various advantages over the UCT algorithm, and has demonstrated better performance on the game of NoGo. The H-MCTS algorithm \cite{huct} performs selection by the SHOT algorithm for nodes that are near to the root and the UCT algorithm for deeper nodes. H-MCTS has also shown superiority over the UCT in games such as $8\times 8$ Amazons and $8\times 8$ AtariGo. Applications of the KL-UCB \cite{klucb} and Thompson sampling \cite{thompson} to MCTS have also been investigated and produced some interesting results\cite{appmab}. 

The improved UCB algorithm \cite{iucb} is a modification of the UCB algorithm, and it has been shown that the improved UCB algorithm has a tighter regret upper bound than the UCB algorithm. In this research, we will explore the possibility of applying the improved UCB algorithm to MCTS. However, some characteristics of the improved UCB algorithm may not be desirable for a direct application to MCTS. Therefore, we have made some modifications to the improved UCB algorithm, making it more suitable for the task of game tree search. We will demonstrate the impact and implications of the modifications we have made on the improved UCB algorithm in an empirical study under the conventional multi-armed bandit problem setting. We will introduce the {\it Mi-UCT} algorithm, which is the application of the modified improved UCB algorithm to MCTS. We will demonstrate the performance of the Mi-UCB algorithm on the game of $9\times 9$ Go and $9\times 9$ NoGo, which has shown to outperform the plain UCT when given a small number of playouts, and roughly on the same level when more playouts are given.

\input{iucb.tex}

\section{Applying Modified Improved UCB Algorithm to Trees}

In this section we will first introduce the improved UCB algorithm. We will then proceed to make some modifications to the improved UCB algorithm, and finally show how to apply the modified algorithm to Monte-Carlo Tree Search.

\subsection{Improved UCB Algorithm}

In the multi-armed bandit problem (MAB), a player is faced with a $K$-armed bandit, and the player can decide to pull one of the arms at each play. The bandit will produce a reward $r\in [0,1]$ according to the arm that has been pulled. The distribution of the reward of each arm is unknown to the player. The objective of the player is to maximize the total amount of reward over $T$ plays. Bandit algorithms are policies that the player can follow to achieve this goal. Equivalent to maximizing the total expected reward, bandit algorithms aim to minimize the cumulative regret, which is defined as
\begin{center}
	$R_t = \sum^{T}_{t=1}r^* - r_{I_t}$,
\end{center}   
where $r^*$ is the expected mean reward of the optimal arm, and $r_{I_t}$ is the received reward when the player chooses to play arm $I_t\in K$ at play $t\in T$. If a bandit algorithm can restrict the cumulative regret to the order of $O(\log T)$, it is said to be optimal \cite{optimal}. The UCB algorithm \cite{ucb}, which is used in the UCT algorithm \cite{uct}, is an optimal algorithm which restricts the cumulative regret to $O(\frac{K\log(T)}{\Delta})$, where $\Delta$ is the difference of expected reward between a suboptimal arm and the optimal arm. The improved UCB algorithm \cite{iucb} is a modification of the UCB algorithm, and it can further restrict the growth of the cumulative regret to the order of $O(\frac{K\log(T\Delta^2)}{\Delta})$. 

The improved UCB algorithm, shown in Algorithm \ref{I-UCB}, essentially maintains a candidate set $B_m$ of potential optimal arms, and then proceeds to systematically eliminate arms which are estimated to be suboptimal from that set. A predetermined number of total plays $T$ is given to the algorithm, and the plays are further divided into $\lfloor \frac{1}{2}\log_2(\frac{T}{e}) \rfloor$ rounds. Each round consists of three major steps.  In the first step, the algorithm samples each arm that is in the candidate set $n_m = \lceil \frac{2\log(T\Delta_m^2)}{\Delta_m^2}\rceil$ times. Next, the algorithm proceeds to remove the arms whose upper bounds of estimated expected reward are less than the lower bound of the current best arm. The estimated difference $\Delta_m$ is then halved in the final step. After each round, the expected reward of the arm $a_i$ is effectively estimated as 

\begin{center}
	$w_i \pm \sqrt{\frac{\log(T\Delta_m^2)}{2n_m}} = w_i \pm \sqrt{\frac{log(T\Delta_m^2)\cdot \Delta_m^2}{4\log(T\Delta_m^2)}}= w_i \pm \frac{\Delta_m}{2}$,
\end{center}

\noindent where $w_i$ is the current average reward received from arm $a_i$.

In the case when the total number of plays $T$ is not predetermined, the improved UCB algorithm can be run in an episodic manner; a total of $T_0 = 2$ plays is given to algorithm in the initial episode, and the number of plays of subsequent episodes is given by $T_{\ell+1} = T_{\ell}^2$.  

\input{miucb.tex}

\subsection{Modification of the Improved UCB Algorithm}

Various characteristics of the improved UCB algorithm might be problematic for its application to MCTS:

\begin{itemize}
	\item {\bf Early explorations}. The improved UCB algorithm tries to find the optimal arm by {\it the process of  elimination}. Therefore, in order to eliminate suboptimal arms as early as possible, it has the tendency to devote more plays to suboptimal arms in the early stages. This might not be ideal when it comes to MCTS, especially in situations when time and resources are rather restricted, because it may end up spending most of the time exploring irrelevant parts of the game tree, rather than searching deeper into more promising subtrees. 

	\item {\bf Not an anytime algorithm}. The improved UCB algorithm requires the total number of plays to be specified beforehand, and its major properties or theoretical guarantees may not hold if it is stopped prematurely. Since we are considering each node as a single instance of the MAB problem in MCTS, internal nodes which are deeper in the tree are most likely the instances that are prematurely stopped. The ``temporal'' solutions provided by these nodes might be erroneous, and the effect of these errors may be magnified as they propagate upward to the root node. On the other hand, it would be rather expensive to ensure the required conditions are met for the improved UCB algorithms on each node, because the necessary amount of playouts will grow exponentially as the number of expanded node increases.     
\end{itemize}

Therefore, we have made some adjustments to the improved UCB algorithm before applying it to MCTS. 

The modified improved UCB bandit algorithm is shown in Algorithm \ref{Mi-UCB}. The modifications try to retain the major characteristics of the improved UCB algorithm, especially the way the confidence bounds are updated and maintained. Nonetheless, we should note that these modifications will change the algorithm's behaviour, and the theoretical guarantees of the original algorithm may no longer be applicable.

\input{iuct.tex}

\subsubsection{Algorithmic Modifications} We have made two major adjustments to the algorithmic aspect of the improved UCB algorithm:

\begin{enumerate}
	\item {\bf Greedy optimistic sampling}. We only sample the arm that currently has the highest upper bound, rather than sampling every possible arm $n_m$ times. 
	\item {\bf Maintain candidate arm count}. We will only maintain the count of potential optimal arms, instead of maintaining a candidate set.
\end{enumerate}
    
Since we are only sampling the current best arm, we are effectively performing a more aggressive arm elimination; arms that are perceived to be suboptimal are not being sampled. Therefore, there is no longer a need for maintaining a candidate set. 

However, the confidence bound in the improved UCB algorithm for arm $a_i$ is defined as $w_i \pm \sqrt{\frac{\log(T\Delta_m^2)}{2n_m}}$, and the updates of $\Delta_m$ and $n_m$ are both dictated by the number of plays in each round, which is determined by $(|B_m|\cdot n_m)$, i.e., the total number of plays that is needed to sample each arm in the candidate set $B_m$ for $n_m$ times. Therefore, in order to update the confidence bound we will need to maintain the count of potential optimal arms.

The implication of sampling the current best arm is that the guarantee for the estimated bound $w_i\pm \Delta_m$ to hold will be higher than the improved UCB algorithm, because the current best will likely be sampled more or equal to $n_m$ times. This is desirable in game tree search, since it would be more efficient to verify a variation is indeed the principal variation, than trying to identify and verify others are suboptimal. 
 
\subsubsection{Confidence Bound Modification} Since we have modified the algorithm to sample only the current best arm, the confidence bound for the current best arm should be tighter than other arms. Hence, an adjustment to the confidence bound is also needed.

In order to reflect the fact that the current best arm is sampled more than other arms, we have modified the definition of the confidence bound for arm $a_i$ to 

\begin{center}
	$w_i \pm \sqrt{\frac{\log(T\Delta_m^2)\cdot r_i}{2n_m}}$,
\end{center}

\noindent where the factor $r_i = \frac{T}{t_i}$, and $t_i$ is the number of times that the arm has been sampled. The more arm $a_i$ is sampled, the smaller $r_i$ will be, and hence the tighter is the confidence bound. Therefore, the expected reward of arm $a_i$ will be estimated as
\begin{center}
	$w_i \pm \sqrt{\frac{\log(T\Delta_m^2)\cdot r_i}{2n_m}} = w_i \pm \sqrt{\frac{log(T\Delta_m^2)\cdot \Delta_m^2 \cdot r_i}{4\log(T\Delta_m^2)}}= w_i \pm \frac{\Delta_m}{2}\sqrt{r_i} = w_i \pm \frac{\Delta_m}{2}\sqrt{\frac{T}{t_i}}$.
\end{center}

Since it would be more desirable that the total number of plays is not required beforehand, we will run the modified improved UCB algorithm in an episodic fashion when we apply it to MCTS, i.e., assigning a total of $T_0 = 2$ plays to the algorithm in the initial episode, and $T_{\ell+1} = T_{\ell}^2$ plays in the subsequent episodes. After each episode, all the relevant terms in the confidence bound, such as $\Delta_m$ and $n_m$, will be re-initialized, and hence information from previous episodes will be lost. Therefore, in order to ``share'' information across episodes, we will not re-initialize $r_i$ after each episode.

\subsection{Modified Improved UCB applied to Trees (Mi-UCT)}

We will now introduce the application of the modified improved UCB algorithm to Monte-Carlo Tree Search, or the {\it Mi-UCT} algorithm. The details of the Mi-UCT algorithm are shown in Algorithm \ref{Mi-UCT}.

The Mi-UCT algorithm adopts the same game tree expansion paradigm as the UCT algorithm, that is, the game tree is expanded over a number of iterations, and each iteration consists of four steps: {\it selection}, {\it expansion}, {\it simulation}, and {\it backpropagation} \cite{uct}. The difference is that the tree policy is replaced by the modified improved UCB algorithm. The modified improved UCB on each node is run in an episodic manner; a total of $T_0 = 2$ plays to the algorithm in the initial episode, and $T_{\ell+1} = T_{\ell}^2$ plays in the subsequent episodes.

The Mi-UCT algorithm keeps track of when $N.\Delta$ should be updated and the starting point of a new episode by using the variables $N.deltaUpdate$ and $N.T$, respectively.  When the number of playouts $N.t$ of the node $N$ reaches the updating deadline $N.deltaUpdate$, the algorithm halves the current estimated regret $N.\Delta$ and calculates the next deadline for halving $N.\Delta$. The variable $N.T$ marks the starting point of a new episode. Hence, when $N.t$ reaches $N.T$, the related variables $N.\Delta$ and $N.armCount$ are re-initialized, and the starting point $N.T$ of the next episode, along with the new $N.deltaUpdate$ are calculated. 
  
\section{Experimental Results}

We will first examine how the various modifications we have made to the improved UCB algorithm affect its performance on the multi-armed bandit problem. Next, we will demonstrate the performance of the Mi-UCT algorithm against the plain UCT algorithm on the game of $9\times 9$ Go and $9 \times 9$ NoGo. 

\subsection{Performance on Multi-armed Bandits Problem}

The experimental settings follow the multi-armed bandit testbed that is specified in \cite{book}. The results are averaged over $2000$ randomly generated $K$-armed bandit tasks. We have set $K=60$ to simulate more closely the conditions in which bandit algorithms will face when they are applied in MCTS for games that have a middle-high branching factor. The reward distribution of each bandit is a normal (Gaussian) distribution with the mean $w_i$, $i\in K$, and variance $1$. The mean $w_i$ of each bandit of every generated $K$-armed bandit task was randomly selected according to a normal distribution with mean $0$ and variance $1$. 

\begin{figure}
\caption{Cumulative Regret of Various Modifications on Improved UCB Algorithm}
\label{fig:CRIUCB}
\centering
	\makebox[\textwidth][c]{\includegraphics[width=\textwidth]{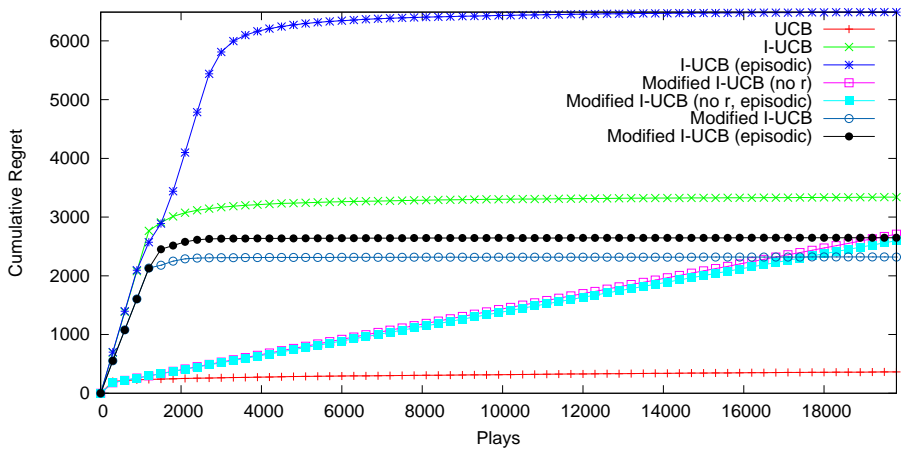}}
\end{figure}

\begin{figure}
\caption{Optimal Arm Percentage of of Various Modifications on Improved UCB Algorithm}
\label{fig:OPIUCB}
\centering
	\makebox[\textwidth][c]{\includegraphics[width=\textwidth]{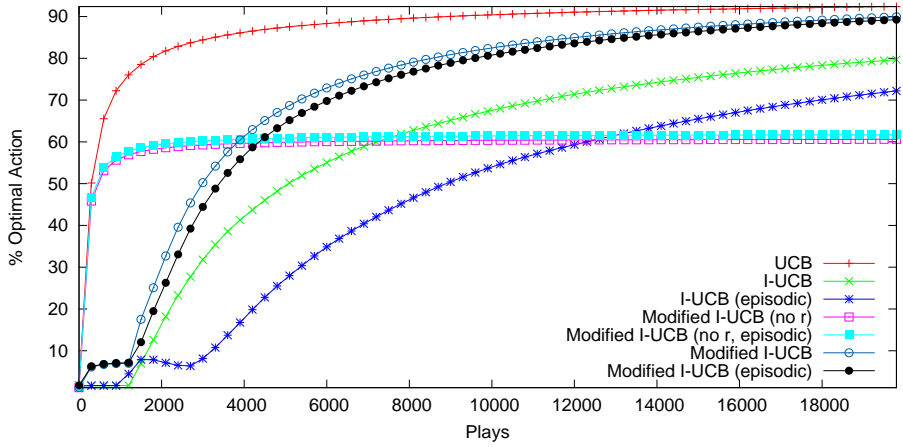}}%
\end{figure}

The cumulative regret and optimal action percentage are shown in Figure \ref{fig:CRIUCB} and Figure \ref{fig:OPIUCB}, respectively. The various results correspond to different algorithms as follows: 
\begin{itemize}
	\item {\bf UCB}: the UCB algorithm.
	\item {\bf I-UCB}: the improved UCB algorithm. 
	\item {\bf I-UCB (episodic)}: the improved UCB algorithm ran episodically.
	\item {\bf Modified I-UCB (no $r$)}: only algorithmic modifications on the improved UCB algorithm.
	\item {\bf Modified I-UCB (no $r$, episodic)}: only algorithmic modifications on the improved UCB algorithm ran episodically.
	\item {\bf Modified I-UCB}: both algorithmic and confidence bound modifications on the improved UCB algorithm. 
	\item {\bf Modified I-UCB (episodic)}: both algorithmic and confidence bound modifications on the improved UCB algorithm ran episodically.
\end{itemize} 

Contrary to theoretical analysis, we are surprised to observe the original improved UCB, both I-UCB and I-UCB(episodic), produced the worst cumulative regret. However, their optimal action percentages are increasing at a very rapid rate, and are likely to overtake the UCB algorithm if more plays are given. This suggests that the improved UCB algorithm does indeed devote more plays to exploration in the early stages. 

The ``slack'' in the curves of the algorithms that were run episodically are the points when a new episode begins. Since the confidence bounds are essentially re-initialized after every episode, effectively extra explorations are performed. Therefore, there were extra penalties on the performance, and it can be clearly observed in the cumulative regret.

We can further see that by making only the algorithmic modification, to give Modified I-UCB (no $r$) and Modified I-UCB(no $r$, episodic), the optimal action percentage increases very rapidly, but it eventually plateaued and stuck to suboptimal arms. Their cumulative regret also increased linearly instead of logarithmically. 

However, by adding the factor $r_i$ to the confidence bound, the optimal action percentage increases rapidly and might even overtake the UCB algorithm if more plays are given. Although the optimal action percentage of the modified improved UCB, both Modified I-UCB and Modified I-UCB (episodic), are rapidly catching up with that of the UCB algorithm; there is still a significant gap between their cumulative regret.       

\subsection{Performance of Mi-UCT against Plain UCT on $9\times 9$ Go}

We will demonstrate the performance of the Mi-UCT algorithm against the plain UCT algorithm on the game of Go played on a $9\times 9$ board. 

For an effective comparison of the two algorithms, no performance enhancing heuristics were applied. The simulations are all pure random simulations without any patterns or simulation policies. A total of $1000$ games were played for each constant $C$ setting of the UCT algorithm, each taking turns to play Black. The total number of playouts was fixed to $1000$, $3000$, and $5000$ for both algorithms. 

The results are shown in Table \ref{tbl:go}. It can be observed that the performance of the Mi-UCT algorithm is quite stable against various constant $C$ settings of the plain UCT algorithm, and is roughly on the same level. The Mi-UCT algorithm seems to have better performance when only $1000$ playouts are given, but slightly deteriorates when more playouts are available.

\input{goresult.tex}

\subsection{Performance of Mi-UCT against Plain UCT on $9\times 9$ NoGo}

We will demonstrate the performance of the the Mi-UCT algorithm against the plain UCT algorithm on the game of NoGo played on a $9\times 9$ board. NoGo is a misere version of the game of Go, in which the first player that has no legal moves other than capturing the opponent's stone loses. 

All the simulations are all pure random simulations, and no extra heuristics or simulation policies were applied. A total of $1000$ games were played for each constant $C$ setting of the UCT algorithm, each taking turns to play Black. The total number of playouts was fixed to $1000$, $3000$, and $5000$ for both algorithms. 

The results are shown in Table \ref{tbl:nogo}. We can observe that the Mi-UCT algorithm significantly dominates the plain UCT algorithm when only $1000$ playouts were given, and the performance deteriorates rapidly when more playouts are available, although it is still roughly on the same level as the plain UCT algorithm.

The results on both $9\times 9$ Go and  $9\times 9$ NoGo suggest that the performance of the Mi-UCT algorithm is comparable to that of the plain UCT algorithm, but scalability seems poorer. Since the proposed modified improved UCB algorithm essentially estimates the expected reward of each bandit by $w_i + \frac{\Delta_m}{2}\sqrt{r_i}$, where $r_i = \sqrt{\frac{T}{t_i}}$, the exploration term converges slower than the of the UCB algorithm, and hence more   exploration might be needed for the modified improved UCB confidence bounds to converge to a ``good-enough'' estimate value; this might be the reason why Mi-UCT algorithm has poor scalability. Therefore, we might able to overcome this problem by trying other definitions for $r_i$. 

\input{nogoresult.tex}

\section{Conclusion}

The improved UCB algorithm is a modification of the UCB algorithm, and has a better regret upper bound than the UCB algorithm. Various characteristics of the improved UCB algorithm, such as early exploration and not being an anytime algorithm, are not ideal for a direct application to MCTS. Therefore, we have made some modifications to the improved UCB algorithm, making it more suitable for the task of game tree search. We have investigated the impact and implications of each modification through an empirical study under the conventional multi-armed bandit problem setting. 

The Mi-UCT algorithm is the application of the modified improved UCB algorithm applied to Monte-Carlo Tree Search.  We have demonstrated that it outperforms the plain UCT algorithm on both games of $9\times 9$ Go and $9\times 9$ NoGo when only a small number of playouts are given, and on comparable level with increased playouts. One possible way of improving the scalability would be trying other definition of $r_i$ in the modified improved UCB confidence bounds.

It would also be interesting to investigate the possibility of enhancing the performance of the Mi-UCT algorithm by combining it with commonly used heuristics \cite{survey} or develop new heuristics that are unique to the Mi-UCT algorithm. Finally, since the modifications we have made essentially changed the behaviour of the original algorithm, investigation into the theoretical properties of our modified improved UCB algorithm may provide further insight into the relation between bandit algorithms and Monte-Carlo Tree Search.

\end{document}

%% file: iucb.tex
\begin{algorithm}
\caption{The Improved UCB Algorithm \cite{iucb}}
\label{I-UCB}
\begin{algorithmic}
\State {\bf Input:} A set of arms $A$, total number of trials $T$
\State {\bf Initialization}: Expected regret $\Delta_0 \gets 1$, a set of candidates arms $B_0\gets A$
\For{rounds $m =0, 1, \cdots,\lfloor \frac{1}{2}\log_2 \frac{T}{e} \rfloor$ }
	\State
	\State {\bf (1) Arm Selection}:
		  \For{all arms $a_i\in B_m$}
		  	\For{$n_m = \lceil \frac{2\log(T\Delta_m^2)}{\Delta^2}\rceil$ times}
		  		\State sample the arm $a_i$ and update its average reward $w_i$ 
		  	\EndFor
		  \EndFor
	 \State	
	 \State {\bf (2) Arm Elimination:}
	 	  \State $a_{max} \gets$ \Call{MaximumRewardArm}{$B_m$}
	 	  \For{all arms $a_i\in B_m$}
	 	  	\If{($w_i + \sqrt{\frac{\log(T\Delta^2)}{2n_m}}) < (w_{max} - \sqrt{\frac{\log(T\Delta^2)}{2n_m}}$) }
	 	  	\State remove $a_i$ from $B_m$
	 	  	\EndIf 
	 	  \EndFor
	 \State
	 \State{\bf (3) Update $\Delta_m$}
	 \State $\Delta_{m+1} = \frac{\Delta_m}{2}$
\EndFor  
\end{algorithmic}
\end{algorithm}

%% file: miucb.tex
\begin{algorithm}
\caption{Modified Improved UCB Algorithm}
\label{Mi-UCB}
\begin{algorithmic}
\State {\bf Input:} A set of arms $A$, total number of trials $T$
\State {\bf Initialization}: Expected regret $\Delta_0 \gets 1$, arm count $N_m \gets |A|$, plays till $\Delta_k$ update $T_{\Delta_0} \gets n_0 \cdot N_m$, where $n_0 \gets \lceil \frac{2\log(T\Delta_0^2)}{\Delta_0^2}\rceil$, number of times arm $a_i\in A$ has been sampled $t_i \gets 0$.
\State
\For{rounds $m =0, 1, \cdots T$ }
	\State
	\State {\bf (1)Sample Best Arm}:
		  	\State  $a_{max} \gets \argmax_{i \in |A|} (w_i + \sqrt{\frac{\log(T\Delta_k^2)\cdot r_i}{2n_k}})$, where $r_i = \frac{T}{t_i}$
		  	\State $w_{max} \gets $ \Call{UpdateMaxWinRate}{$A$}
		  	\State $t_i\gets t_i+1$
	 \State	
	 \State {\bf (2) Arm Count Update:}
	 	  \For{all arms $a_i$}
	 	  	\If{($w_i + \sqrt{\frac{\log(T\Delta_k^2)}{2n_k}}) < (w_{max} - \sqrt{\frac{\log(T\Delta_k^2)}{2n_k}}$) }
	 	  	\State $N_m \gets N_m - 1$
	 	  	\EndIf 
	 	  \EndFor
	 \State
	 \State{\bf (3) Update $\Delta_k$ when Deadline $T_{\Delta_k}$ is Reached}
	 \If{$m \geq T_{\Delta_k}$}
	 	\State $\Delta_{k+1} = \frac{\Delta_k}{2}$
	 	\State $n_{k+1} \gets \lceil \frac{2\log(T\Delta_{k+1}^2)}{\Delta_{k+1}^2}\rceil$
	 	\State $T_{\Delta_{k+1}} \gets m + (n_{k+1} \cdot N_m)$ 
	 	\State $k \gets k + 1$
	 \EndIf
\EndFor  
\end{algorithmic}
\end{algorithm}

%% file: iuct.tex
\begin{algorithm}
\caption{Modified Improved UCB Algorithm applied to Trees (Mi-UCT)}
\label{Mi-UCT}
\begin{algorithmic}
\Function {Mi-UCT}{Node $N$}
\State $best_{ucb} \gets -\infty$
\For{all child nodes $n_i$ of $N$}
	\If{$n_i.t = 0$}
		\State $n_i.ucb \gets \infty$
	\Else
	    \State $r_i\gets N.episodeUpdate / n_i.t $
		\State $n_i.ucb \gets n.w + \sqrt{\frac{\log(N.T \times N.\Delta^2)\times r_i}{2N.k}}$
	\EndIf
	\If{$best_{ucb} \leq n_i.ucb$}
		\State $best_{ucb}\gets n_i.ucb$
		\State $n_{best}\gets n_i$
	\EndIf
\EndFor
\State
\If{$n_{best}.times = 0$}
	\State $result \gets $\Call{RandomSimulation}{($n_{best}$)}
\Else
	\State {\bf if} $n_{best}$ is not yet expanded {\bf then} \Call{NodeExpansion}{$(n_{best})$}  
	\State $result \gets $ \Call{Mi-UCT}{$(n_{best})$}
\EndIf
\State 
\State $N.w \gets (N.w \times N.t + result) / (N.t + 1)$
\State $N.t \gets N.t + 1$
\State
\If{$N.t \geq N.T$}
	\State $N.\Delta \gets 1$
	\State $N.T \gets N.t + N.T \times N.T$
	\State $N.armCount \gets$ Total number of child nodes
	\State $N.k\gets \lceil \frac{2\log(N.T \times N.\Delta^2)}{N.\Delta^2}\rceil$
	\State $N.deltaUpdate \gets N.t + N.k \times N.armCount$
\EndIf
\State
\If{$N.t \geq N.deltaUpdate$}
	\For{all child nodes $n_i$ of $N$}
		\If{($n_i.w + \sqrt{\frac{\log(N.T\times N.\Delta^2)}{2n.k}}) < (N.w - \sqrt{\frac{\log(N.T\times N.\Delta^2)}{2n.k}}$) }
			\State $N.armCount \gets N.armCount - 1$
	 	\EndIf 
	\EndFor
	\State
	\State $N.\Delta \gets \frac{N.\Delta}{2}$
	\State $N.k \gets \lceil \frac{2\log(N.T\times N.\Delta^2)}{N.\Delta^2}\rceil$
	\State $N.deltaUpdate \gets N.t + N.k \times N.armCount$
\EndIf
\State 
\Return $result$
\EndFunction 
\State
\Function{NodeExpansion}{Node $N$}
\State $N.\Delta \gets 1$
\State $N.T \gets 2$
\State $N.armCount \gets$ Total number of child nodes
\State $N.k\gets \lceil \frac{2\log(N.t \times N.\Delta^2)}{N.\Delta^2}\rceil$
\State $N.deltaUpdate\gets N.k \times N.armCount$
\EndFunction
 
\end{algorithmic}
\end{algorithm}

%% file: goresult.tex
\begin{table}
\caption{Win rate of Mi-UCT against plain UCT on $9\times 9$ Go}
\label{tbl:go}
\centering
\begin{tabular}{c|ccccccccc}
     constant C & 0.1 & 0.2 & 0.3 & 0.4 & 0.5 & 0.6 & 0.7 & 0.8 & 0.9 \\
  \hline
   1000 playouts & 57.1\% & 55.2\%  & 57.5\% & 52.2\% & 58.6\% & 58.4\% & 55.8\% & 55.3\% & 54.5\% \\
   3000 playouts & 50.8\% & 50.9\%  & 50.3\% & 52.2\% & 52.2\% & 54.4\% & 56.5\% & 56.0\% & 54.1\% \\
   5000 playouts & 54.3\% & 54.2\%  & 52.4\% & 51.0\% & 52.4\% & 57.5\% & 54.9\% & 56.1\% & 55.3\% \\
\end{tabular}
\end{table}

%% file: nogoresult.tex
\begin{table}
\caption{Win rate of Mi-UCT against plain UCT on $9\times 9$ NoGo}
\label{tbl:nogo}
\centering
\begin{tabular}{c|ccccccccc}
     constant C & 0.1 & 0.2 & 0.3 & 0.4 & 0.5 & 0.6 & 0.7 & 0.8 & 0.9 \\
  \hline
   1000 playouts & 58.5\% & 56.1\%  & 61.4\% & 56.7\% & 57.4\% & 58.4\% & 59.6\% & 56.9\% & 57.8\% \\
   3000 playouts & 50.3\% & 51.4\%  & 53.1\% & 51.0\% & 49.6\% & 54.4\% & 56.0\% & 54.2\% & 53.9\% \\
   5000 playouts & 45.8\% & 48.8\%  & 48.5\% & 49.6\% & 55.1\% & 51.3\% & 51.3\% & 55.0\% & 52.7\% \\
\end{tabular}
\end{table}

%% file: typeinst.bbl
\begin{thebibliography}{4}

\bibitem{optimal} Lai, T.L., Robbins, H.: Asymptotically efficient adaptive allocation rules. Advances in applied mathematics 6 (1): 4 (1985)

\bibitem{iucb} Auer, P., Ortner, R.: UCB revisited: Improved regret bounds for the stochastic multi-armed bandit problem. Periodica Mathematica Hungarica 61, pp. 1-2 (2010)

\bibitem{uct} Kocsis, L., Szepesv\'{a}ri, C.: Bandit Based Monte-carlo Planning. Proceedings of the 17th European Conference on Machine Learning (ECML'06), pp. 282-293 (2006)

\bibitem{ucb} Auer, P., Cesa-Bianchi, N., Fischer, P.: Finite-time Analysis of the Multiarmed Bandit Problem.Machine Learning 47, Issue 2-3, pp. 235-256 (2002)

\bibitem{book} Sutton, R. S., Barto, A. G.: Reinforcement Learning: An Introduction. MIT Press, Cambridge, MA, (1998)

\bibitem{survey} Browne, C.B., Powley, E.,  Whitehouse, D., Lucas, S.M., Cowling, P.I., Rohlfshagen, P., Tavener, S., Perez, D., Samothrakis, S., Colton, S.: A Survey of Monte Carlo Tree Search Methods. IEEE Trans. Comp. Intell. AI Games 4(1), pp. 1-43 (2012)

\bibitem{simple} Tolpin,D., Shimony, S.E.: MCTS Based on Simple Regret. Proceedings of the 26th AAAI Conference on Artificial Intelligence, pp. 570-576 (2012)

\bibitem{shot} Cazenave, T.: Sequential Halving applied to Trees. IEEE Trans. Comp. Intell. AI Games volPP, no.99, pp.1-1 (2014) 

\bibitem{huct} Pepels, T., Cazenave, T., Winands, M.H.M., Lanctot, M.: Minimizing Simple and Cumulative Regret in Monte-Carlo Tree Search. Proceedings of Computer Games Workshop at the 21st European Conference on Artificial Intelligence (2014)

\bibitem{appmab} Imagawa, T.,Kaneko, T.: Applying Multi Armed Bandit Algorithms to MCTS and Those Analysis. Proceedings of the 19th Game Programming Workshop (GPW-14), pp.145-150 (2014)

\bibitem{sh} Karnin, Z., Koren, T., Oren, S.: Almost Optimal Exploration in Multi-Armed Bandits. Proceedings of the 30th International Conference on Machine Learning (ICML'13), pp.1238-1246 (2013)

\bibitem{klucb} Garivier, A., Capp´e, A.:The KL-UCB algorithm for bounded stochastic bandits and beyond. Proceedings of 24th Annual Conference on Learning Theory (COLT '11), pp.359-376 (2011)

\bibitem{thompson} Kaufmann, E., Korda, N., Munos, R.: Thompson Sampling: An Asymptotically Optimal Finite-Time Analysis. Proceedings of 23rd Algorithmic Learning Theory (ALT'12), pp.199-213 (2012)

\end{thebibliography}
